# Visual Understanding via Multi-Feature Shared Learning with Global Consistency

Lei Zhang, *Member, IEEE*, and David Zhang, *Fellow, IEEE*

*Abstract*—Image/video data is usually represented with multiple visual features. Fusion of multi-source information for establishing the attributes has been widely recognized. Multi-feature visual recognition has recently received much attention in multimedia applications. This paper studies visual understanding via a newly proposed $\ell_2$-norm based multi-feature shared learning framework, which can simultaneously learn a global label matrix and multiple sub-classifiers with the labeled multi-feature data. Additionally, a group graph manifold regularizer composed of the Laplacian and Hessian graph is proposed for better preserving the manifold structure of each feature, such that the label prediction power is much improved through the semi-supervised learning with global label consistency. For convenience, we call the proposed approach Global-Label-Consistent Classifier (GLCC). The merits of the proposed method include: 1) the manifold structure information of each feature is exploited in learning, resulting in a more faithful classification owing to the global label consistency; 2) a group graph manifold regularizer based on the Laplacian and Hessian regularization is constructed; 3) an efficient alternative optimization method is introduced as a fast solver owing to the convex sub-problems. Experiments on several benchmark visual datasets for multimedia understanding, such as the 17-category Oxford Flower dataset, the challenging 101-category Caltech dataset, the YouTube & Consumer Videos dataset and the large-scale NUS-WIDE dataset, demonstrate that the proposed approach compares favorably with the state-of-the-art algorithms. An extensive experiment on the deep convolutional activation features also show the effectiveness of the proposed approach. The code is available on http://www.escience.cn/people/lei/index.html

## I. Introduction

Multiple modalities, multiple views and multiple features are usually used to represent the multimedia contents and images. For example, given a face image, its visual content can be represented with several kinds of weak modalities such as the left and right periocular, mouth and nose regions [4]; given a video frame, its visual content can be represented by different feature types such as the histogram, SIFT, HSV, etc. [9]. With multi-feature representation, how to exploit the rich *structural* information of each feature in modeling is a challenging task in multimedia analysis.

At the early stage, information fusion can be divided into three levels: feature level, score level and decision level. Feature-level fusion was demonstrated to be more effective for recognition than the score-level and decision-level fusions [16]. Feature concatenation is recognized as a prevalent fusion method in pattern recognition [18], [19]. However, it is less effective in multimedia content analysis, due to that the visual features are often independent or heterogeneous [17]. In particular, the complexity of data analysis becomes high if one simply concatenates feature vectors into a high-dimensional feature vector. For those reasons, multi-view learning concept has been developed by the researchers in machine learning community. One popular work was the two-view based support vector machine (SVM-2k) [11], [21], [22], which learned one SVM model with two views of the data. Another popular work was multiple kernel learning (MKL) [10], [20], which aimed at integrating the information of multiple features together by combining multiple kernels with appropriate weights in learning. Additionally, the concept of multi-modal joint learning is also involved in dictionary learning and sparse representation. Some representative methods under the framework of multi-dictionary learning can be referred to as [25], [26], [27], [28], and [29], which have been proposed for visual recognition such as face, digit, action and object recognition. The reported results demonstrate that learning multiple discriminative dictionaries or visually correlated dictionaries can effectively improve the recognition performance with a reconstruction-error based classifier [24]. Recently, several multi-modal joint sparse representation methods were also developed for pattern recognition applications. For example, in [3], a multi-task joint sparse representation classifier (MTJSRC) was proposed for visual classification, in which the group sparsity was used to combine multiple features. In [4], a kernel space based multi-modal sparse model was proposed for robust face recognition. In [30], a joint dynamic sparse representation model was proposed for object recognition. In [48], a very efficient multi-task feature selection model (FSSI) with low-rank constraint was proposed for multimedia analysis.

This work was supported by National Natural Science Foundation of China (No. 61401048), the 2013 Innovative Team Construction Project of Chongqing Universities, the Hong Kong Scholar Program (No.XJ2013044) and China Postdoctoral Science Foundation (No. 2014M550457).
• L. Zhang is with the College of Communication Engineering, Chongqing University, and the Department of Computing, The Hong Kong Polytechnic University, Hong Kong. (e-mail: leizhang@cqu.edu.cn).
• D. Zhang is with the Department of Computing, The Hong Kong Polytechnic University, Hong Kong (e-mail: csdzhang@comp.polyu.edu.hk).

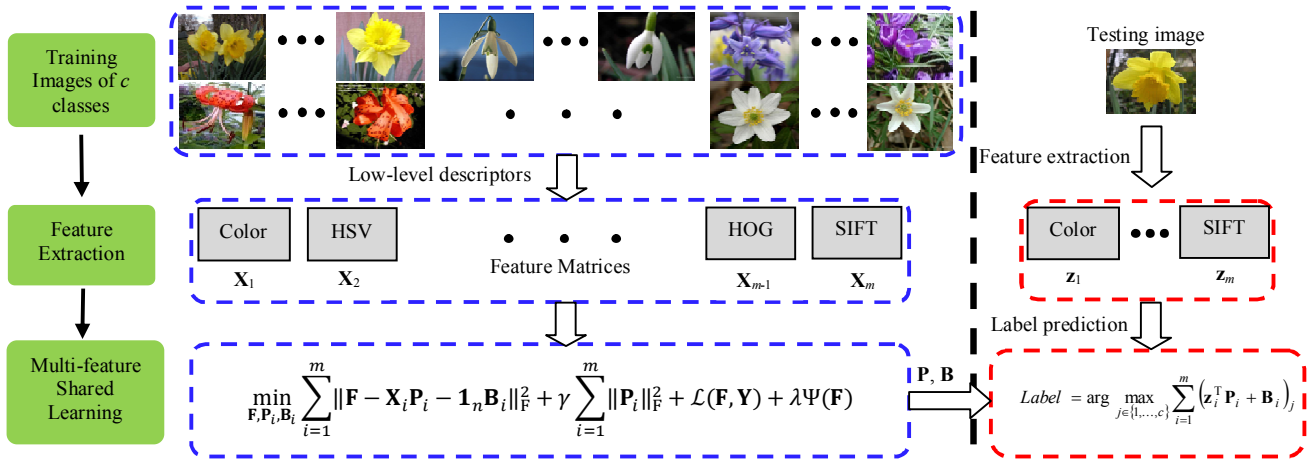

Fig. 1. Overview of the proposed framework. In the left part (the training phase), the proposed algorithm exploits a multi-feature shared learning over $m$ potential visual features $\mathbf{X}_i \in \mathbb{R}^{n \times d_i}, i = 1, \cdots, m$ of the training images. In the right part (the testing phase), a joint decision function with the learned classifier parameters $\mathbf{P}_i \in \mathbb{R}^{d_i \times c}$ and $\mathbf{B}_i^T \in \mathbb{R}^c$ is computed based on the extracted $m$ visual features $\mathbf{z}_i \in \mathbb{R}^{d_i}, i = 1, \cdots, m$ from the testing image. *Notations*: $\mathbf{F}$ denotes the predicted label matrix, $\mathbf{Y}$ denotes the given label matrix, $\mathcal{L}(\mathbf{F}, \mathbf{Y})$ is the loss function, $\Psi(\mathbf{F})$ is the group graph manifold regularizer, $n$ is the number of training samples, $d_i$ is the number of dimensions of the $i$-th feature, $c$ denotes the number of classes, $\gamma$ and $\lambda$ denote the regularization coefficients. More details about the proposed minimization model can be referred to as the Section III.

Motivated by these multi-task/multi-modal joint learning models, we present a multi-feature learning concept that aims at exploiting the complementary structural information of features. Although the joint learning concept has been proved to be effective in classification, it still faces with a dilemma of insufficient labeled data that are costly and expensive to label in hand in real-world applications. In this paper, we focus on the semi-supervised learning framework for pursuit of further improvement of the multi-feature learning capability when the label information of the training data is insufficient. It is known that the Laplacian graph based manifold regularization is the mainstream in semi-supervised learning, owing to its better exploration of the intrinsic data geometry. However, the Laplacian graph has been identified to be biased towards a constant function when there are only a few labeled data, due to the constant null space and the weakly preserved local topology [5]. Further, the Hessian graph has been proved to have a better extrapolating power than the Laplacian graph from two aspects: 1) it has a richer null space; 2) it can exploit the intrinsic local geometry of the manifold structure very well [5].

For better exploiting the manifold structure of each feature in semi-supervised learning, motivated by the spirit of joint learning concepts discussed above, we target at proposing a multi-feature shared learning framework based on the Hessian and Laplacian graphs. Additionally, we also expect that the underlying feature correlation and complementary structural information among multiple features can be exploited for simultaneously learning multiple predictors during the shared learning process. For this reason, we advocate learning multiple features with global consistency based on a weighted group graph manifold regularizer, resulting in a more faithful classification performance when only a few labeled data is available. The concept of global consistency in this paper is useful for cooperative learning of multiple features and manifolds with the global objective (label). It is worth noting that there is no explicit mapping matrix in the manifold regression during the testing process, which also motivates us to present an explicit and semi-supervised classifier based on the proposed group graph manifold regularization.

With the considerations of those above concerns, a multi-feature shared learning framework with global consistency based on the weighted Laplacian and Hessian graph is proposed in this paper for visual understanding. In terms of the essential idea in this proposal, the proposed approach is nominated as Global-Label-Consistent Classifier (GLCC) for discussion. The merits of this paper are as follows.

- Correlation and complementary information of multiple features are exploited in the shared learning model with the global consistency.
- For better exploiting the manifold structure of the few labeled training data, a group graph regularizer based on the Hessian and Laplacian graphs is developed for preserving the global label consistency.
- Considering that there is no explicit predictor in the manifold regression, an explicit joint classifier is learned by minimizing the least-square alike loss function with the global label prediction.
- In the proposed method, a $\ell_2$-norm based global classifier with a very efficient alternative optimization is presented.

The overview of the proposed GLCC framework is illustrated in Fig.1. The experiments have been conducted on several benchmark visual datasets, such as the Oxford flower 17 dataset[1] from [12], the Caltech 101 dataset[2] from [14], the YouTube & Consumer video dataset[3] from [45], the large-scale NUS-WIDE web image dataset[4] from [53], and an extensive convolutional neural net (CNN) based deep convolutional activation feature set (DeCAF) of object images from 4 domains [54, 56]. All experiments demonstrate that our GLCC outperforms many existing multi-feature and semi-supervised learning methods. In particular, GLCC can also work well with the deep features.

---
[1] http://www.robots.ox.ac.uk/~vgg/data/flowers/17/index.html
[2] http://www.robots.ox.ac.uk/~vgg/software/MKL/
[3] http://vc.sce.ntu.edu.sg/transfer learning domain adaptation/domain adaptation home.html
[4] http://lms.comp.nus.edu.sg/research/NUS-WIDE.html

The rest of this paper is organized as follows. In Section II, we review the most related work in visual recognition, graph based semi-supervised learning and graph based multi-view learning. The proposed GLCC framework with the formulation and optimization algorithm is described in Section III. The experiments on several benchmark datasets are discussed in Section IV. The convergence and computational time are briefly discussed in section V. Section VI concludes this paper.

## II. RELATED WORK

In this section, we briefly review the current prevailing approaches on visual recognition, graph based semi-supervised learning and graph based multi-view learning.

### A. Visual Recognition

A number of methods have been developed for face recognition, gender recognition, age estimation, scene categories and object recognition in computer vision community. The bag-of-features (BoF) model is a popular image categorization method, but it discards the spatial order of local descriptors and degrades the descriptive power of an image. For this reason, Lazebnik et al. [2] proposed a spatial pyramid matching (SPM) beyond the bags of features for natural scene and object recognition. Yang et al. [40] proposed a linear SPM based on sparse coding (ScSPM) for visual classification with significant improvement. Gehler et al. [1] proposed two feature combination methods such as the average kernel support vector machine (AK-SVM) and the product kernel support vector machine (PK-SVM). Additionally, multiple kernel learning (MKL) [23], [37], [38], the column generation boosting (CG-Boost) [13] and the linear programming boosting (LP-B and LP-β) [15] have also been proposed for object recognition. However, there is a common flaw of these methods that the computational cost is too large. Recently, Yuan et al. [3] proposed a multi-task joint sparse representation (MTJSRC) based on $\ell_{1,2}$ mixed-norm for visual classification, which shows a better performance by comparing with several sparse dictionary learning methods including [24], [25], [26], [27] and [28]. Zhang et al. [30] proposed a multi-observation joint dynamic sparse representation for visual recognition with a competitive performance.

### B. Graph based Semi-supervised Learning

Semi-supervised learning has been widely deployed in the recognition tasks, due to the fact that training a small amount of labeled data is prone to overfitting on one hand, and the manual labeling process of a large number of unlabeled data is tedious and time-consuming on the other hand. Some good examples of semi-supervise learning are presented. For example, Belkin et al. [7] proposed a Laplacian graph manifold based semi-supervised learning framework, in which a manifold assumption that the manifold structure information of the unlabeled data can be preserved was defined. The consistency assumption implies that the nearby data points or the data points on the same cluster/manifold are likely to have the same label. Note that the cluster assumption is local while the manifold assumption is global. Belkin et al. [41] also proposed a manifold regularization framework for semi-supervised learning, in which the Laplacian regularized least square and the Laplacian support vector machine were discussed. Zhou et al. [31] proposed a graph based semi-supervised method (LGC) for learning the local and global consistency through the graph regularization framework. Ma et al. [51] proposed a semi-supervised feature selection algorithm (SFSS) for multimedia analysis based on the Laplacian graph and the $l_{2,1}$-norm regularization. In the graph manifold based algorithms, a consensus is that the affinity information is used to classify the unlabeled data. Additionally, the Laplacian eighenmap based manifold learning was usually used for dimension reduction and graph embedding in single view/modality [6], [32], [33], [34]. With these manifold methods discussed above, the Laplacian graph in single-view is the mainstream of semi-supervised learning, but it has been identified that the solution is biased towards a constant with weak extrapolating power [5]. Instead, the Hessian graph was proved to have a good extrapolating power in manifold regularization. In this paper, we have a full consideration of the Hessian graph manifold regularizer in our proposed approach.

### C. Graph based Multi-view Learning

Multi-view graph manifold regression has been reported in recent years. For example, Wang et al. [8] proposed a subspace sharing based semi-supervised multi-feature method for action recognition, in which both the global and local consistency were considered in classifier training. Tong et al. [42] proposed a graph based multi-modality learning method with the linear and sequential fusion schemes, but the mapping function in the objective function is implicit. Xia et al. [35] proposed a graph Laplacian based multi-view spectral embedding (MSE) for dimension reduction, which solves an eigenvalue problem in optimization. Wu et al. [43] proposed a sparse multi-modal dictionary learning model with the Laplacian hyper-graph regularizer. Yang et al. [9] proposed a multi-feature Laplacian graph based hierarchical semi-supervised regression (MLHR) method for multimedia analysis and achieved better performance in video concept annotation. In this paper, motivated by the multi-view graph based learning concept, an idea of multi-feature shared learning is introduced.

## III. MULTI-FEATURE SHARED LEARNING FRAMEWORK

In this section, the proposed Global-Label-Consistent Classifier (GLCC) with the model formulation, optimization, training algorithm and recognition is presented.

### A. Notations

Assume that there are $n$ training samples of $c$ classes. Denote $\mathbf{X}_i = [\mathbf{x}_1, \mathbf{x}_2, \cdots, \mathbf{x}_n]^T \in \mathbb{R}^{n \times d_i}$ as the training set of the $i$-th feature modality ($i$=1,…,$m$), $\mathbf{Y} = [\mathbf{y}_1, \mathbf{y}_2, \cdots, \mathbf{y}_n]^T \in \mathbb{R}^{n \times c}$ as the global label matrix of the training samples, and $\mathbf{F}=[\mathbf{F}_1,\mathbf{F}_2,\cdots,\mathbf{F}_n]^T\in\mathbb{R}^{n\times c}$ as the predicted label matrix of the training data, where $d_i$ denotes the dimension of the $i$-th feature and $m$ denotes the number of features. In this paper, $\|\cdot\|_F$ denotes the Frobenius norm, $\|\cdot\|_2$ denotes the $\ell_2$-norm, and $\mathbf{Tr}(\cdot)$ denotes the trace operator. Given a sample vector $\mathbf{x}_i$, $y_{ij}=1$ if $\mathbf{x}_i$ belongs to the $j$-th class, and $y_{ij}=0$, otherwise. The learned sub-classifier for the $i$-th feature is defined as $\mathbf{P}_i\in\mathbb{R}^{d_i\times c}$ plus a bias $\mathbf{B}_i^T\in\mathbb{R}^c$. The Laplacian and Hessian graph matrix are represented as $\mathcal{L}$ and $\mathbf{\Omega}$, respectively.

*B. Formulation of GLCC*

Semi-supervised learning is generally with the manifold assumption that the nearby data points are more likely to have the same labels. In the graph based manifold learning, label consistency is preserved in the manifold structure of data geometry. Motivated by [8], [35], [41], [42], and [43], the proposed GLCC is generally formulated as follows

$$\min_{\mathbf{F},\mathbf{P}_i,\mathbf{B}_i}\sum_{i=1}^m\|\mathbf{F}-\mathbf{X}_i\mathbf{P}_i-\mathbf{1}_n\mathbf{B}_i\|_F^2+\gamma\sum_{i=1}^m\|\mathbf{P}_i\|_F^2+\mathcal{L}(\mathbf{F},\mathbf{Y})+\lambda\Psi(\mathbf{F}) \quad (1)$$

where $\gamma$ and $\lambda$ are the positive trade-off parameters, $\mathbf{1}_n\in\mathbb{R}^n$ is a full one vector, $\mathbf{F}$ is the predicted label matrix, $\mathcal{L}(\cdot)$ is the loss function, and $\Psi(\cdot)$ is the graph manifold regularization term.

For convenience, let $Q(\mathbf{F})=\mathcal{L}(\mathbf{F},\mathbf{Y})+\lambda\Psi(\mathbf{F})$, then the graph based manifold regression model can then be written as

$$Q(\mathbf{F})=\mathcal{L}(\mathbf{F},\mathbf{Y})+\lambda\Psi(\mathbf{F})$$
$$=\sum_i\ell oss(f(\mathbf{x}_i),\mathbf{y}_i)+\lambda\sum_i\sum_j\mathcal{A}_{i,j}\|f(\mathbf{x}_i)-f(\mathbf{x}_j)\|_2^2 \quad (2)$$

where $\ell oss(\cdot)$ denotes the least-square loss function, $\lambda$ is the regularization parameter ($\lambda>0$), and $\mathcal{A}$ denotes the adjacency matrix whose entries are defined as

$$\mathcal{A}_{i,j}=\begin{cases}1,\text{ if }\mathbf{x}_i\in\mathcal{N}_k(\mathbf{x}_j)\text{ or }\mathbf{x}_j\in\mathcal{N}_k(\mathbf{x}_i)\\0,\quad\quad\quad\quad\quad\quad\quad\quad\text{otherwise}\end{cases} \quad (3)$$

where $\mathcal{N}_k(\mathbf{x}_j)$ denotes the local set consisting of the $k$-nearest neighbors of $\mathbf{x}_j$.

The least-square loss function term in (2) can be written as

$$\sum_i\ell oss(f(\mathbf{x}_i),\mathbf{y}_i)=\sum_i w_i\|f(\mathbf{x}_i)-\mathbf{y}_i\|_2^2$$
$$=\mathbf{Tr}(\mathbf{F}-\mathbf{Y})^T\mathbf{W}(\mathbf{F}-\mathbf{Y}) \quad (4)$$

where $\mathbf{W}$ is a diagonal matrix with the entries $W_{ii}$ defined as follows: for semi-supervised use, $W_{ii}$ is set as a large value (e.g. $10^{10}$) if the $i$-th sample is labeled, and 0 otherwise.

The second term in (2) is the manifold structure preservation term for global label consistency. Specifically, the Laplacian graph is used in part to preserve the label information in the manifold built on the training data. It can be written in trace-form as

$$\sum_i\sum_j\mathcal{A}_{i,j}\|f(\mathbf{x}_i)-f(\mathbf{x}_j)\|_2^2=\mathbf{Tr}(2\mathbf{F}^T\mathcal{L}\mathbf{F}) \quad (5)$$

where $\mathcal{D}$ is a diagonal matrix with the entries $\mathcal{D}_{ii}=\sum_j\mathcal{A}_{i,j}$ and $\mathcal{L}=\mathcal{D}-\mathcal{A}$ is the Laplacian graph matrix.

As denoted in [5], the Laplacian graph based semi-supervised learning suffers from the fact that the solution is biased towards a constant with weak extrapolating power if only a few labeled points are available. Instead, the second-order Hessian energy regularizer was proved to have a better extrapolation capability than the Laplacian graph. Specifically, the total estimated Hessian energy is shown as

$$\widehat{\mathbf{S}}_{\text{Hess}}(f)=\langle\mathbf{F},\mathbf{\Omega}\mathbf{F}\rangle=\mathbf{Tr}(\mathbf{F}^T\mathbf{\Omega}\mathbf{F}) \quad (6)$$

where $\mathbf{\Omega}$ is the Hessian energy matrix and it is sparse since each data point is only associated with its neighbors. The details of the Hessian energy estimation are shown in Appendix A.

For exploiting the advantages of both the Laplacian and Hessian graph regularizers, the proposed manifold regularization with a group graph regularizer is represented as

$$\min_{\mathbf{F}}Q(\mathbf{F}) \quad (7)$$

where $Q(\mathbf{F})$ in terms of (2) and (4) can be re-written as

$$Q(\mathbf{F})=\mathbf{Tr}(\mathbf{F}-\mathbf{Y})^T\mathbf{W}(\mathbf{F}-\mathbf{Y})+\lambda\cdot\mathbf{Tr}(\mathbf{F}^T(\mathcal{L}+\mathbf{\Omega})\mathbf{F}) \quad (8)$$

However, the representation of $Q(\mathbf{F})$ in (8) is in single feature. In this paper, the multi-feature concept is exploited. Therefore, the objective function $Q(\mathbf{F})$ with $m$ features can be formulated as

$$Q(\mathbf{F})=\mathbf{Tr}(\mathbf{F}-\mathbf{Y})^T\mathbf{W}(\mathbf{F}-\mathbf{Y})+\lambda\cdot\mathbf{Tr}(\mathbf{F}^T(\sum_{i=1}^m\alpha_i^r\mathcal{L}^{(i)}+\sum_{i=1}^m\beta_i^r\mathbf{\Omega}^{(i)})\mathbf{F}) \quad (9)$$

where $\alpha_i^r$ and $\beta_i^r$ ($0<\alpha,\beta<1;r>1$) denote the contribution coefficients of the Laplacian matrix $\mathcal{L}^{(i)}$ and the Hessian energy matrix $\mathbf{\Omega}^{(i)}$ w.r.t. the $i$-th feature, and the equality constraint of $\alpha_i$ and $\beta_i$, i.e. $\sum_{i=1}^m\alpha_i=\sum_{i=1}^m\beta_i=1$ is required for better exploring the contribution of each feature. In this paper, for convenience, we define $\mathbf{G}=\sum_{i=1}^m\alpha_i^r\mathcal{L}^{(i)}+\sum_{i=1}^m\beta_i^r\mathbf{\Omega}^{(i)}$ as the group graph regularizer. Note that the setting of $r>1$ is for better exploiting the complementary information of multiple features and avoiding the trivial solution with only the best feature considered (e.g. $\alpha_i=1$). For this reason, $\alpha_i^r$ and $\beta_i^r$ instead of $\alpha_i$ and $\beta_i$ are used in this paper.

In the proposed group graph based manifold regularization model (7), we observe that there is no explicit classifier to predict the label matrix $\mathbf{F}$. We therefore propose to learn the multi-feature based global classifiers $\mathbf{P}=\{\mathbf{P}_1,\cdots,\mathbf{P}_m\}$ and $\mathbf{B}=\{\mathbf{B}_1,\cdots,\mathbf{B}_m\}$ for predicting $\mathbf{F}$, as formulated as (1). Suppose that $\mathbf{X}_i=[\mathbf{x}_1^i,\mathbf{x}_2^i,\cdots,\mathbf{x}_n^i]^T$ is the training set of the $i$-th feature with $n$ samples, the multi-feature based global classifier model can be written as

$$\{\mathbf{P},\mathbf{B}\}$$
$$=\arg\min_{\mathbf{P},\mathbf{B}}\sum_{i=1}^m\sum_{j=1}^n\|\mathbf{y}_j-\mathbf{x}_j^i\mathbf{P}_i-\mathbf{B}_i\|_F^2+\gamma\sum_{i=1}^m\|\mathbf{P}_i\|_F^2$$
$$=\arg\min_{\mathbf{P},\mathbf{B}}\sum_{i=1}^m\|\mathbf{Y}-\mathbf{X}_i\mathbf{P}_i-\mathbf{1}_n\mathbf{B}_i\|_F^2+\gamma\sum_{i=1}^m\|\mathbf{P}_i\|_F^2 \quad (10)$$

where $\mathbf{1}_n$ denotes a column vector with all ones, $\gamma$ denotes the balance parameter ($0<\gamma<1$), and $\|\mathbf{P}_i\|_F^2$ is used to control the complexity and avoid overfitting.

By combining the group graph based manifold regularization model (7) and the multi-feature global classifier (10) together,

the GLCC is formulated. In summary, the GLCC framework shown in (1) can be finally re-written as follows

$$\min_{\mathbf{F},\mathbf{P},\mathbf{B},\alpha_i,\beta_i} \sum_{i=1}^{m}\|\mathbf{F}-\mathbf{X}_i\mathbf{P}_i-\mathbf{1}_n\mathbf{B}_i\|_F^2 + \gamma \sum_{i=1}^{m}\|\mathbf{P}_i\|_F^2$$
$$+\text{Tr}(\mathbf{F}-\mathbf{Y})^{\text{T}}\mathbf{W}(\mathbf{F}-\mathbf{Y}) + \lambda \cdot \text{Tr}\left(\mathbf{F}^{\text{T}}\left(\sum_{i=1}^{m}\alpha_i^r\mathcal{L}^{(i)} + \sum_{i=1}^{m}\beta_i^r\mathbf{\Omega}^{(i)}\right)\mathbf{F}\right) \quad (11)$$
$$\text{s.t.} \sum_{i=1}^{m}\alpha_i = \sum_{i=1}^{m}\beta_i = 1, 0 < \gamma, \lambda < 1, \alpha_i, \beta_i > 0, r > 1$$

In (11), the first term denotes the multi-feature based global label predictor, the second regularization term is to control the complexity and avoid overfitting in the learning phase, the third term is the least-square alike loss function and the final term is the group graph manifold regularizer for preserving the global consistency and similarity of the labeled data. Parameters $\alpha_i^r$ and $\beta_i^r$ denote the weights of Laplacian and Hessian graph *w.r.t.* the $i^{\text{th}}$ feature, and $r>1$ denotes that it can make full use of the information of all features rather than the best feature (e.g. $\alpha_i=1$, $\beta_j=1$), such that the complementary *structural* information of different features can be fully exploited [36].

*C. Optimization*

From the proposed GLCC framework (11), we observe that the solutions can be easily solved with a very efficient alternative optimization approach.

First, we fix the $\alpha_i = \beta_i = 1/m, \forall i \in (1,\cdots,m)$. The initialized $\mathbf{F}$ can be solved by setting the derivative of the following objective function *w.r.t.* $\mathbf{F}$ to be 0,

$$\min_{\mathbf{F}} \text{Tr}(\mathbf{F}-\mathbf{Y})^{\text{T}}\mathbf{W}(\mathbf{F}-\mathbf{Y}) + \lambda \cdot \text{Tr}\left(\mathbf{F}^{\text{T}}\left(\sum_{i=1}^{m}\alpha_i^r\mathcal{L}^{(i)} + \sum_{i=1}^{m}\beta_i^r\mathbf{\Omega}^{(i)}\right)\mathbf{F}\right) \quad (12)$$

Then, the $\mathbf{F}$ can be initialized as

$$\mathbf{F} = \left(\mathbf{W} + \left(\sum_{i=1}^{m}\alpha_i^r\mathcal{L}^{(i)} + \sum_{i=1}^{m}\beta_i^r\mathbf{\Omega}^{(i)}\right)\right)^{-1}\mathbf{W}\mathbf{Y} \quad (13)$$

After fixing the $\mathbf{F}$, $\alpha_i$ and $\beta_i$, the optimization problem shown in (11) becomes

$$\min_{\mathbf{P}_i,\mathbf{B}_i} \sum_{i=1}^{m}\|\mathbf{F}-\mathbf{X}_i\mathbf{P}_i-\mathbf{1}_n\mathbf{B}_i\|_F^2 + \gamma \sum_{i=1}^{m}\|\mathbf{P}_i\|_F^2 \quad (14)$$

By setting the derivatives of the objective function (14) *w.r.t.* $\mathbf{P}_i$ and $\mathbf{B}_i$ to be 0, respectively, we have

$$\mathbf{P}_i = (\mathbf{X}_i^{\text{T}}\mathbf{X}_i + \gamma\mathbf{I})^{-1}(\mathbf{X}_i^{\text{T}}\mathbf{F} - \mathbf{X}_i^{\text{T}}\mathbf{1}_n\mathbf{B}_i) \quad (15)$$

$$\mathbf{B}_i = (\mathbf{1}_n^{\text{T}}\mathbf{1}_n)^{-1}(\mathbf{1}_n^{\text{T}}\mathbf{F} - \mathbf{1}_n^{\text{T}}\mathbf{X}_i\mathbf{P}_i) \quad (16)$$

where $\mathbf{I}$ is an identity matrix and $\mathbf{1}_n$ is a full one vector. Note that in computing $\mathbf{P}_i$ (15), the $\mathbf{B}_i$ is initialized as zero.

After fixing the $\mathbf{P}_i, \mathbf{B}_i, \alpha_i, \beta_i$, the optimization problem becomes

$$\min_{\mathbf{F}} \sum_{i=1}^{m}\|\mathbf{F}-\mathbf{X}_i\mathbf{P}_i-\mathbf{1}_n\mathbf{B}_i\|_F^2 + \text{Tr}(\mathbf{F}-\mathbf{Y})^{\text{T}}\mathbf{W}(\mathbf{F}-\mathbf{Y}) + \lambda \cdot \text{Tr}\left(\mathbf{F}^{\text{T}}\left(\sum_{i=1}^{m}\alpha_i^r\mathcal{L}^{(i)} + \sum_{i=1}^{m}\beta_i^r\mathbf{\Omega}^{(i)}\right)\mathbf{F}\right) \quad (17)$$

By setting the derivative of the objective function (17) *w.r.t.* $\mathbf{F}$ to be 0, the predicted label matrix $\mathbf{F}$ can be solved as

$$\mathbf{F} = (m\mathbf{I} + \mathbf{W} + \lambda\mathbf{G})^{-1}\left(\sum_{i=1}^{m}(\mathbf{X}_i\mathbf{P}_i + \mathbf{1}_n\mathbf{B}_i) + \mathbf{W}\mathbf{Y}\right) \quad (18)$$

where $\mathbf{G} = \sum_{i=1}^{m}\alpha_i^r\mathcal{L}^{(i)} + \sum_{i=1}^{m}\beta_i^r\mathbf{\Omega}^{(i)}$.

---

**Algorithm 1.** The proposed GLCC method
**Input:**
The training data of $m$ features $\mathbf{X}_i \in \mathbb{R}^{n \times d_i}, i = 1,\cdots,m$;
The training labels $\mathbf{Y} \in \mathbb{R}^{n \times c}$;
The parameters $\lambda$, $\gamma$, and $r$;
**Output:**
The $\mathbf{P}_i \in \mathbb{R}^{d_i \times c}$ and $\mathbf{B}_i \in \mathbb{R}^c, i = 1,\cdots,m$;
**Procedure:**
1. Compute the graph Laplacian matrices $\mathcal{L}^{(i)} \in \mathbb{R}^{n \times n}$;
2. Compute the Hessian energy matrices $\mathbf{\Omega}^{(i)} \in \mathbb{R}^{n \times n}$;
3. Compute the selection matrix $\mathbf{W} \in \mathbb{R}^{n \times n}$;
4. Initialize the $\alpha_i \leftarrow 1/m$, $\beta_i \leftarrow 1/m$, $\mathbf{B}_i \leftarrow \mathbf{0} \in \mathbb{R}^c$;
5. Initialize the $\mathbf{F}$ according to the (13);
6. **While not converged do**
   Compute the $\mathbf{P}_i$ according to the (15);
   Compute the $\mathbf{B}_i$ according to the (16);
   Update the $\mathbf{F}$ according to the (18);
   Update the $\alpha_i$ and $\beta_i$ according to the (22);
   **until** Convergence;
7. Return the $\mathbf{P}_i$ and $\mathbf{B}_i$;

---

After fixing the $\mathbf{F}, \mathbf{P}_i, \mathbf{B}_i$, the optimization of $\alpha_i$ and $\beta_i$ becomes

$$\min_{\alpha_i,\beta_i} \text{Tr}\left[\mathbf{F}^{\text{T}}\left(\sum_{i=1}^{m}\alpha_i^r\mathcal{L}^{(i)} + \sum_{i=1}^{m}\beta_i^r\mathbf{\Omega}^{(i)}\right)\mathbf{F}\right] \quad (19)$$
$$\text{s.t.} \sum_{i=1}^{m}\alpha_i = \sum_{i=1}^{m}\beta_i = 1$$

The Lagrange equation of (19) can be written as

$$Lag(\alpha_i,\beta_i,\mu,\eta) = \text{Tr}\left[\mathbf{F}^{\text{T}}\left(\sum_{i=1}^{m}\alpha_i^r\mathcal{L}^{(i)} + \sum_{i=1}^{m}\beta_i^r\mathbf{\Omega}^{(i)}\right)\mathbf{F}\right] - \mu\left(\sum_{i=1}^{m}\alpha_i - 1\right) - \eta\left(\sum_{i=1}^{m}\beta_i - 1\right) \quad (20)$$

where $\mu$ and $\eta$ denote the Lagrange multiplier coefficients.

By setting the derivative of (20) *w.r.t.* $\alpha_i$, $\beta_i$, $\mu$, $\eta$ to be 0, respectively, we have

$$\begin{cases} r\alpha_i^{r-1}\text{Tr}(\mathbf{F}^{\text{T}}\mathcal{L}^{(i)}\mathbf{F}) - \mu = 0 \\ r\beta_i^{r-1}\text{Tr}(\mathbf{F}^{\text{T}}\mathbf{\Omega}^{(i)}\mathbf{F}) - \eta = 0 \\ \sum_{i=1}^{m}\alpha_i - 1 = 0 \\ \sum_{i=1}^{m}\beta_i - 1 = 0 \end{cases} \quad (21)$$

where the parameters $\alpha_i$ and $\beta_i$ can be solved as follows

$$\begin{cases} \alpha_i = \left(\frac{1}{\text{Tr}(\mathbf{F}^{\text{T}}\mathcal{L}^{(i)}\mathbf{F})}\right)^{1/(r-1)} \Big/ \sum_{i=1}^{m}\left(\frac{1}{\text{Tr}(\mathbf{F}^{\text{T}}\mathcal{L}^{(i)}\mathbf{F})}\right)^{1/(r-1)} \\ \beta_i = \left(\frac{1}{\text{Tr}(\mathbf{F}^{\text{T}}\mathbf{\Omega}^{(i)}\mathbf{F})}\right)^{1/(r-1)} \Big/ \sum_{i=1}^{m}\left(\frac{1}{\text{Tr}(\mathbf{F}^{\text{T}}\mathbf{\Omega}^{(i)}\mathbf{F})}\right)^{1/(r-1)} \end{cases} \quad (22)$$

where the $\mathbf{F}$ is represented as (18). The details of solving the (21) for $\mathbf{\alpha}$ and $\mathbf{\beta}$ are provided in Appendix B.

Consequently, an iterative training procedure for solving the optimization model (11) is summarized in the Algorithm 1. According to the Algorithm 1, we can infer that the objective function of (11) monotonically decreases until convergence. The proofs are given in the following sub-section *E*.

*D. Recognition*

The classifier parameters $\{\mathbf{P}_i\}_{i=1}^{m}$ and $\{\mathbf{B}_i\}_{i=1}^{m}$ can be obtained by using the Algorithm 1 with the training set. In recognition, the label of a given testing image represented with $m$ features $\mathbf{z}_i \in \mathbb{R}^d, i = 1, \ldots, m$ can be calculated as

$$label = \arg \max_{j \in \{1, \cdots, c\}} \left[\sum_{i=1}^{m} (\mathbf{z}_i^T \mathbf{P}_i + \mathbf{B}_i)\right]_j \quad (23)$$

where $\sum_{i=1}^{m}(\mathbf{z}_i^T \mathbf{P}_i + \mathbf{B}_i)$ denotes the output of $c$-dimensional vector. Specifically, the recognition procedure of the proposed GLCC framework is summarized in the Algorithm 2.

*E. Convergence*

In order to prove the convergence behavior of the proposed Algorithm 1, we first provide a lemma as follows.

**Lemma 1**: *For alternative optimization, when update one variable with other variables fixed, that is, update $\mathbf{P}_i^t, \mathbf{B}_i^t, \mathbf{F}^t, \alpha_i^t$, and $\beta_i^t$ ($t$ denotes the iteration index) will not increase the objective function $\mathcal{J}(\cdot)$. Four claims are given:*

*Claim 1.* $\mathcal{J}(\mathbf{P}_i^t, \mathbf{B}_i^t, \mathbf{F}^t, \alpha_i^t, \beta_i^t) \geq \mathcal{J}(\mathbf{P}_i^{t+1}, \mathbf{B}_i^t, \mathbf{F}^t, \alpha_i^t, \beta_i^t)$

*Proof.* After fixing $\mathbf{B}_i, \mathbf{F}, \alpha_i, \beta_i$, the objective function is convex w.r.t. $\mathbf{P}_i$, and a closed form solution (15) can be obtained by setting the derivative w.r.t. $\mathbf{P}_i$ as 0. For this reason, it is clear that $\mathcal{J}(\mathbf{P}_i^t, \mathbf{B}_i^t, \mathbf{F}^t, \alpha_i^t, \beta_i^t) \geq \mathcal{J}(\mathbf{P}_i^{t+1}, \mathbf{B}_i^t, \mathbf{F}^t, \alpha_i^t, \beta_i^t)$.

*Claim 2.* $\mathcal{J}(\mathbf{P}_i^{t+1}, \mathbf{B}_i^t, \mathbf{F}^t, \alpha_i^t, \beta_i^t) \geq \mathcal{J}(\mathbf{P}_i^{t+1}, \mathbf{B}_i^{t+1}, \mathbf{F}^t, \alpha_i^t, \beta_i^t)$

*Proof.* Similar to the proof of *claim 1*, the objective function is convex w.r.t. $\mathbf{B}_i$ after fixing $\mathbf{P}_i, \mathbf{F}, \alpha_i, \beta_i$. Then, we have $\mathcal{J}(\mathbf{P}_i^{t+1}, \mathbf{B}_i^t, \mathbf{F}^t, \alpha_i^t, \beta_i^t) \geq \mathcal{J}(\mathbf{P}_i^{t+1}, \mathbf{B}_i^{t+1}, \mathbf{F}^t, \alpha_i^t, \beta_i^t)$.

*Claim 3.* $\mathcal{J}(\mathbf{P}_i^{t+1}, \mathbf{B}_i^{t+1}, \mathbf{F}^t, \alpha_i^t, \beta_i^t) \geq \mathcal{J}(\mathbf{P}_i^{t+1}, \mathbf{B}_i^{t+1}, \mathbf{F}^{t+1}, \alpha_i^t, \beta_i^t)$

*Proof.* When $\mathbf{P}_i, \mathbf{B}_i, \alpha_i, \beta_i$ are fixed, the optimization problem (17) is convex w.r.t. $\mathbf{F}$. By setting the derivative of the objective function (17) w.r.t. $\mathbf{F}$ to be 0, the solution (18) of $\mathbf{F}$ can decrease the objective function. *Claim 3* is proven.

*Claim 4.* $\mathcal{J}(\mathbf{P}_i^{t+1}, \mathbf{B}_i^{t+1}, \mathbf{F}^{t+1}, \alpha_i^t, \beta_i^t) \geq \mathcal{J}(\mathbf{P}_i^{t+1}, \mathbf{B}_i^{t+1}, \mathbf{F}^{t+1}, \alpha_i^{t+1}, \beta_i^{t+1})$

*Proof.* As can be seen from (21), with $\mathbf{P}_i$, $\mathbf{B}_i$, and $\mathbf{F}$ fixed, the update rule of $\alpha_i$ and $\beta_i$ are obtained by setting the derivatives of objective function (20) w.r.t. $\alpha_i$ and $\beta_i$ to be 0. The second-order derivatives w.r.t. $\alpha_i$ and $\beta_i$ are as follows

$$\frac{d^2 L(\alpha_i, \beta_i, \mu, \eta)}{d\alpha_i^2} = r(r-1)\alpha_i^{r-2} \mathbf{Tr}(\mathbf{F}^T \boldsymbol{\mathcal{L}}^{(i)} \mathbf{F}) > 0; \ r > 1, \alpha_i > 0$$

$$\frac{d^2 L(\alpha_i, \beta_i, \mu, \eta)}{d\beta_i^2} = r(r-1)\beta_i^{r-2} \mathbf{Tr}(\mathbf{F}^T \boldsymbol{\Omega}^{(i)} \mathbf{F}) > 0; \ r > 1, \beta_i > 0$$

Since both the second-order derivatives are positive, the update rule (22) of $\alpha_i$ and $\beta_i$ can be guaranteed to decrease the objective function (20). *Claim 4* is proven.

Further, the convergence of the proposed iteration method in Algorithm 1 is summarized in the following theorem.

**Theorem 1**: *The objective function (11) monotonically decreases until convergence after several iterations by using Algorithm 1.*

*Proof.* Suppose the updated $\mathbf{P}_i^t$, $\mathbf{B}_i^t$, $\mathbf{F}^t$, $\alpha_i^t$ and $\beta_i^t$ are $\mathbf{P}_i^{t+1}$, $\mathbf{B}_i^{t+1}$, $\mathbf{F}^{t+1}$, $\alpha_i^{t+1}$, and $\beta_i^{t+1}$, respectively. According to *claim 1*, *claim 2*, *claim 3* and *claim 4* presented in **lemma 1**, we observe that

---

**Algorithm 2.** Recognition of GLCC framework
**Input:**
Training set $\{\mathbf{X}_i\}_{i=1}^{m}$, training labels $\mathbf{Y}$, and one test sample $\{\mathbf{z}_i\}_{i=1}^{m}$ of $m$ features;
**Procedure:**
Obtain $\{\mathbf{P}_i\}_{i=1}^{m}$ and $\{\mathbf{B}_i\}_{i=1}^{m}$ by solving model (11) using the proposed **Algorithm 1** on the training set.
**Output:**
$label \leftarrow \arg \max_{j \in \{1, \cdots, c\}} \left[\sum_{i=1}^{m} (\mathbf{z}_i^T \mathbf{P}_i + \mathbf{B}_i)\right]_j$

---

$$\mathcal{J}(\mathbf{P}_i^t, \mathbf{B}_i^t, \mathbf{F}^t, \alpha_i^t, \beta_i^t) \geq \mathcal{J}(\mathbf{P}_i^{t+1}, \mathbf{B}_i^t, \mathbf{F}^t, \alpha_i^t, \beta_i^t)$$
$$\geq \mathcal{J}(\mathbf{P}_i^{t+1}, \mathbf{B}_i^{t+1}, \mathbf{F}^t, \alpha_i^t, \beta_i^t)$$
$$\geq \mathcal{J}(\mathbf{P}_i^{t+1}, \mathbf{B}_i^{t+1}, \mathbf{F}^{t+1}, \alpha_i^t, \beta_i^t)$$
$$\geq \mathcal{J}(\mathbf{P}_i^{t+1}, \mathbf{B}_i^{t+1}, \mathbf{F}^{t+1}, \alpha_i^{t+1}, \beta_i^{t+1})$$

Then **Theorem 1** is proven.

*F. Computational Complexity*

We now briefly analyze the computational complexity of the proposed GLCC method, which involves $T$ iterations and $m$ kinds of features. The time complexity of computing the Laplacian and Hessian matrices is $O(mn^3)$. In the learning phase, each iteration involves four update steps in Algorithm 1, and the time complexity for all iterations is $O(m^2 ndT)$. Hence, the total computational complexity of our method is $O(mn^3) + O(m^2 ndT)$. Note that computation of the Laplacian and Hessian matrices for all features is implemented before iterations, such that the total computational complexity can be reduced. Additionally, the computational time of the proposed method for different datasets in experiments is presented in Sections IV, and further discussed in Section V.

*G. Remarks*

From the level of approach, the in-depth motivation behind the proposal is that the multi-feature shared learning framework with global consistency is capable of exploiting the *correlation* and complementary *structural* information of multiple features, such that the manifold structure of individual feature can be well preserved and considered. In general, *structural* information of an image is preserved after feature description, which is independent from other information (e.g. luminance). In this paper, the manifold embedding structure is considered.

First, in GLCC, the manifold structure of the $i$-th feature is represented with the Laplacian graph $\boldsymbol{\mathcal{L}}^{(i)}$ and the Hessian graph $\boldsymbol{\Omega}^{(i)}$. In order to exploit the complementary information of $m$ features, the weights $\alpha_i$ and $\beta_i$ of the two manifold graphs are learned in optimization, respectively. Therefore, the group graph regularizer $\Psi(\mathbf{F})$ is proposed for global consistency preservation. Second, for semi-supervised learning with only a few labeled data available, a least-square alike loss function $\mathcal{L}(\mathbf{F}, \mathbf{Y})$ is proposed by introducing a diagonal selection matrix $\mathbf{W}$. Third, the concept of global consistency is proposed for cooperative learning among multiple features and manifolds, such that multiple sub-predictors $\{\mathbf{P}_i\}_{i=1}^{m}$ and bias $\{\mathbf{B}_i\}_{i=1}^{m}$ have been easily learned with the global prediction $\mathbf{F}$.

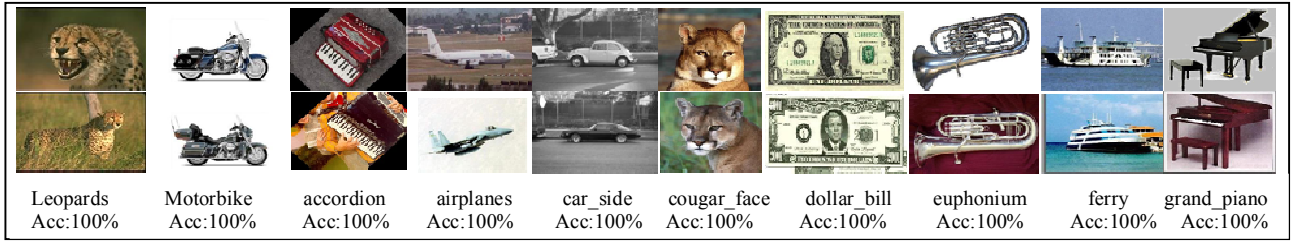

Fig. 2. Example images the first 10 classes (2 images per class) of 100% recognition accuracy with our GLCC on the Caltech 101 dataset.

## IV. EXPERIMENTS

In this section, the experiments are conducted on the Oxford Flowers 17 dataset, the Caltech 101 dataset, the YouTube & Consumer Videos dataset and the large-scale NUS-WIDE dataset for multimedia understanding. Additionally, we have also conducted an extensive experiment on the convolutional neural net (CNN) based deep features for object recognition.

### A. Datasets, Features and Experimental Setup

*Oxford Flowers 17 Dataset:* The Flower 17 dataset consists of 17 species and 1360 images with 80 images per category. The authors in [44] provided seven $\chi^2$-distance matrices as features, such as clustered HSV, HOG, SIFT on the foreground internal region (SIFTint), SIFT on the foreground boundary (SIFTbdy) and three matrices derived from color, shape and texture vocabularies, respectively. Three predefined splits of training (40 images per class), validation (20 images per class) and testing (20 images per class) are considered. We strictly follow the experimental settings in [1], [3], [13], [15], [37] and [38] that the same three predefined train/test splits are used in all methods for fair comparison. This dataset is used to validate the proposed GLCC for 17-class flower recognition task.

*Caltech 101 Dataset:* The Caltech 101 dataset is a challenging object recognition dataset, which contains 9144 images of 101 object categories as well as a background class. For fair comparison, we strictly follow the experimental settings stated by the developer of the dataset. Four kinds of kernel matrices extracted using the MKL code package [39], such as geometric blur (GB), Phow-gray ($L$=0, 1, 2), Phow-color ($L$=0, 1, 2) and SSIM ($L$=0, 1, 2), have been used in this paper. Note that $L$ is the spatial-pyramid level. For all algorithms, 15 training images per category and 15 testing images per category according to the three predefined training/testing splits [3] are discussed in experiments. The example images of the first 10 classes of 100% recognition accuracy with our GLCC are described in Fig. 2.

*YouTube & Consumer Videos Dataset:* The dataset contains 195 consumer videos (target domain) and 906 YouTube videos (auxiliary domain) of six events, such as birthday, picnic, parade, show, sports and wedding. The dataset was developed in [45] for domain adaptation tasks. We strictly follow the experimental setting in [45] for all methods. Specifically, 906 loosely labeled YouTube videos in the source domain and 18 videos (i.e. three samples per event) in the target domain are selected as the labeled training data. The remaining consumer videos in target domain are used as the testing data. Five random splits of the training and testing data from the target domain are experimented and evaluated by using the means and standard deviations of the MAPs (mean average precision). The videos are described by the SIFT ($L$=0 and $L$=1) features and the space-time (ST with $L$=0 and $L$=1) features [45].

*NUS-WIDE Dataset:* the dataset is a large-scale web image set including 269648 real-world scene and object images of 81 concepts, such as the airport, animals, clouds, buildings, etc. In this dataset, six types of descriptors were used to extract low level features, including the 144-D color correlogram (CORR), 73-D edge direction histogram (EDH), 128-D wavelet texture (WT), 225-D block-wise color moments (CM), 64-D color histogram (CH) and 500-D bag of words (BOG) feature based on SIFT. In our experiments, the first three types of visual features such as the CORR, EDH and WT are considered. We randomly select 3000 samples from the dataset for model training, and the remaining data are used for model testing. Different percentages of the labeled data in the training data, such as 10%, 30%, 50%, 70% and 90% are discussed. The mean average precision (MAP) is evaluated. We run the procedure 10 times, and the mean MAPs are reported.

*CNN-Features:* the CNN-features denote the deep representations of object images with a well-trained CNN. In this paper, the Deep Convolutional Activation Feature (DeCAF) [54] is considered. The CNN network was trained on the challenging ImageNet-1000, and the network structure is the same as the proposed CNN in [55], which includes 5 convolutional layers and 3 fully-connected layers. The well-trained network parameters are used for deep representation of the well-known *4DA* dataset including the Caltech (1123), Amazon (958), Webcam (295) and Dslr (157) domains with 10 object classes [56]. Note that the numeric in each bracket denotes the number of samples in each domain. The outputs of the 6-th ($f_6$) and 7-th ($f_7$) fully-connected layers of CNN are recognized as two types of features in this paper. The dimension of features in $f_6$ and $f_7$ is 4096.

### B. Parameter Settings

In GLCC model, there are two regularization parameters $\lambda$ and $\gamma$. The parameters $\lambda$ and $\gamma$ are tuned from the set $\{10^{-4}, 10^{-2}, 1, 10^2, 10^4\}$ throughout the experiments, and the best results are reported. The maximum iteration number is set as 5. The parameter sensitivity analysis is discussed in the subsection *I*.

### C. Experimental Results on the Flower 17 Dataset

The comparison experiments on the Flower 17 dataset are discussed in two parts. First, we compare with the baseline and state-of-the-art results of 11 methods reported in the previous work. Second, in order to further demonstrate the effectiveness of the proposed model, we also compare with four challenging methods such as FSNM [52], FSSI [48], SFSS [51], and MLHR

TABLE I
BRIEF COMPARISON OF DIFFERENT METHODS

| Method | Supervised | Semi-supervised | Single-feature | Multi-feature |
|---|---|---|---|---|
| FSNM | √ | | √ | |
| FSSI | √ | | | √ |
| SFSS | | √ | √ | |
| MLHR | | √ | | √ |
| GLCC | | √ | | √ |

TABLE II
COMPARISONS WITH STATE-OF-THE-ART METHODS BY FEATURE
COMBINATION ON OXFORD FLOWER 17 DATASET

| Methods | Accuracy (%) | Time (s) |
|---|---|---|
| NS Combination | 83.2±2.1 | - |
| SRC Combination | 85.9±2.2 | - |
| AK-SVM [1] | 84.9±1.2 | 2 |
| PK-SVM [1] | 85.5±1.2 | 10 |
| MKL(SILP) [38] | 85.2±1.5 | 97 |
| MKL(simple) [37] | 85.2±1.5 | 152 |
| CG-Boost [13] | 84.8±2.2 | 1.2e3 |
| LP-β [15] | 85.5±3.0 | 80 |
| LPBoost [15] | 85.4±2.4 | 98 |
| FDDL [26] | 86.7±1.3 | 1.9e3 |
| KMTJSRC [3] | 86.8±1.5 | 16 |
| FSNM [52] | 85.9±0.7 | 24 |
| FSSI [48] | 86.9±2.4 | 12 |
| SFSS [51] | 85.6±1.0 | 282 |
| MLHR [9] | 86.7±2.4 | 20 |
| **GLCC** | **87.2±2.2** | 14 |

[9] that have close relation with the proposed GLCC. The brief descriptions of these methods are shown in Table I. In experiments, we have tuned the parameters of each method, and report their best results. The results of all methods are described in Table II, in which the average recognition accuracies and the total computational time (s) are provided. We see that the proposed GLCC obtains the highest recognition accuracy of 87.2%, which outperforms the state-of-the-art accuracy (86.8%) obtained by using the previous KMTJSRC [3]. The GLCC is also better than the multi-feature and semi-supervised learning methods such as FSSI [48] and MLHR [9]. Additionally, the total computational time of GLCC is 14 seconds, and it is still competitive by comparing to the state-of-the-arts.

For deep discussions of FSNM, FSSI, SFSS, MLHR and GLCC, five percentages such as 10%, 30%, 50%, 70% and 90% of the training data are determined as labeled data, respectively, with the remaining data as unlabeled data. Under different percentages, we observe the performance variation of different methods with increasing number of labeled training data. The recognition accuracies of the five methods on the Flower 17 data are shown in Fig.3(a). The bar plot clearly shows that for different percentages of labeled data, the proposed method always outperforms other methods. The experiment on the Flower 17 preliminarily demonstrates the effectiveness and efficiency of our method.

D. *Experimental Results on Caltech 101 Data*

This data shows a more challenging task than the Flower 17 data, owing to the 101 categories. First, we report the results of the baseline and state-of-the-arts proposed in the previous work, such as the NS, SRC, MKL [39], LPBoost [15] and KMTJSRC [3] in Table III. We can observe that our proposed GLCC achieves the best recognition (73.5%) and outperforms the state-of-the-art KMTJSRC (71.0%). Second, the multi-feature and semi-supervised methods such as FSNM, FSSI, SFSS, and MLHR are tested on this dataset, and their best results after parameter tuning are also reported in Table III. We see that the FSSI obtains the second better accuracy 73.2% which is 0.3% lower than our GLCC and the MLHR ranks the third. Notably, we observe that FSNM and SFSS achieve the worst recognition performance. This may show the importance of multi-feature learning in improving the classification performance. The computational time for each method is shown in Table III. From the perspective of accuracy and computation, our GLCC is more effective and computationally efficient than others.

Additionally, the performance variation with different percentages of labeled training data is described in Fig.3(b). It is clear that the proposed GLCC outperforms other methods. In particular, the FNSM and SFSS without utilizing multi-feature learning show the worst recognition performance.

For this popular object dataset, we have to mention the result of CNN based deep learning. As shown in [57], the recognition accuracy on the Caltech 101 data with 15 objects per class as training is 83.8%, where the features are represented with a pre-trained CNN network on ImageNet. However, if the CNN is directly trained on the Caltech 101 data without using any extra data, the recognition accuracy only achieves 46.5%, which shows the ineffectiveness of training a large CNN on such a small dataset [57]. Due to the difference in the training protocol and data, it is not appropriate to compare with deep learning on the considered datasets in this paper.

E. *Experimental Results for Video Event Recognition*

By following the experimental protocol of the YouTube & Consumer videos dataset, all methods are compared in three cases: a) classifiers learned based on SIFT features with $L=0$ and $L=1$; b) classifiers learned based on ST features with $L=0$ and $L=1$; c) classifiers learned based on both SIFT and ST features with $L=0$ and $L=1$. The results are shown in Table IV.

First, we compare our GLCC with several baseline methods such as SVM-T, SVM-AT, MKL, adaptive SVM (A-SVM) [46] and FR [47]. Notably, SVM-AT denotes that the labeled training data are from both auxiliary domain and target domain, while SVM-T denotes that the labeled training data are only from target domain. From Table IV, we observe that the proposed method achieves the highest MAP 44.9% in average which outperforms the best baseline result of MKL. It is worth noting that the domain adaptation methods reported in [45] for this dataset are not compared because our method does not belong to a transfer learning framework.

Second, by comparing with FSNM, FSSI, SFSS and MLHR, we see that MLHR obtains the second best result 43.7% in average and 1.2% lower than our GLCC. Moreover, the result of SIFT features (i.e. case (a)) is much better than that of ST features (i.e. case (b)). The multi-feature learning of both SIFT and ST features (i.e. case (c)) shows comparative results as well as case (a). Additionally, as can be seen from case (c), the multi-feature learning methods such as FSSI, MLHR and the GLCC show significantly higher precision than FSNM and SFSS methods. The computational time shown in Table IV demonstrates the efficiency of our GLCC.

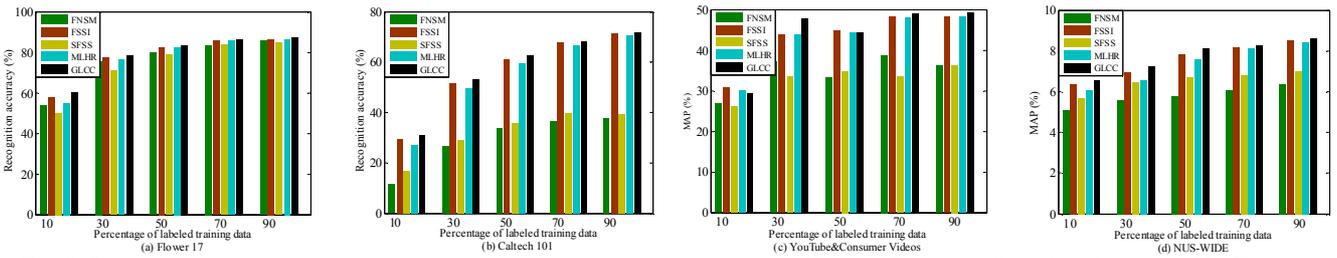

Fig. 3. Performance variants *w.r.t.* different percentages of labeled training data on Flower 17 data (a), Caltech 101 data (b), YouTube & Consumer videos data (c) and NUS-WIDE data (d).

TABLE III
RECOGNITION ACCURACY ON THE CALTECH-101 DATASET

| Method | NS | SRC | MKL [39] | LPBoost [15] | KMTJSRC [3] | FSNM [52] | FSSI [48] | SFSS [51] | MLHR [9] | **GLCC** |
|---|---|---|---|---|---|---|---|---|---|---|
| Accuracy (%) | 51.7±0.8 | 69.2±0.7 | 70.0±0.4 | 70.7±0.4 | 71.0±0.3 | 41.4±0.7 | 73.2±0.2 | 42.00 | 72.4±0.3 | **73.5±0.2** |
| Time (s) | - | - | 1380 | 2135 | 155 | 57.9 | 28.7 | 147.3 | 47.0 | 33.2 |

TABLE IV
MEANS AND STANDARD DEVIATIONS (%) OF MAPs OVER SIX EVENTS FOR ALL METHODS IN THREE CASES

| Method | SVM-T | SVM-AT | FR [47] | A-SVM [46] | MKL [39] | FSNM [52] | FSSI [48] | SFSS [51] | MLHR [9] | **GLCC** |
|---|---|---|---|---|---|---|---|---|---|---|
| MAP-(a) | 42.3±5.5 | **53.9±5.6** | 50.0±5.6 | 38.4±7.9 | 47.2±2.6 | 48.2±3.2 | 49.6±4.0 | 43.2±2.5 | 48.7±4.3 | 49.7±3.9 |
| MAP-(b) | 32.6±2.1 | 24.7±2.2 | 28.4±2.6 | 25.0±1.3 | 35.3±1.6 | 33.3±1.0 | 32.3±0.8 | 32.2±0.5 | 34.5±0.7 | **35.7±0.8** |
| MAP-(c) | 42.0±4.9 | 36.2±3.4 | 44.1±3.6 | 32.4±5.0 | 46.9±2.5 | 39.2±2.8 | 47.5±2.7 | 42.4±2.2 | 47.9±0.9 | **49.4±1.8** |
| Average | 39.0±4.2 | 38.3±3.7 | 40.8±3.9 | 31.9±4.7 | 43.2±2.2 | 40.3±2.4 | 43.2±2.5 | 39.3±1.7 | 43.7±2.0 | **44.9±2.2** |
| Time (s) | 18.0 | 34.4 | 70.3 | 80.5 | 98.1 | 22.6 | 25.4 | 35.3 | 42.1 | 34.0 |

Third, the performance variation with different percentages of labeled training data on this dataset is described in Fig.3(c). As before, our method outperforms other algorithms, except for the cases of 10% and 50%.

### F. Experimental Results on NUS-WIDE Dataset

For the NUS-WIDE data, we compare our GLCC with the existing multi-feature learning and semi-supervised methods, such as FSNM, FSSI, SFSS and MLHR. By training the models on 3000 training samples, the MAPs of the test data for different methods are reported in Table V. We can observe that our GLCC outperforms other methods in recognition ability. The computational time also shows the efficiency of the proposed GLCC. Additionally, the performance variation with 10%, 30%, 50%, 70% and 90% of labeled training data is shown in Fig.3(d). We can clearly observe that our GLCC outperform other methods.

### G. Experimental Results on CNN-Features

The extensive experiments on the CNN features of object datasets from Amazon, Caltech, Webcam, Dslr domains are discussed. By following the experimental setting in [56], 20, 8, 8, and 8 samples per class are randomly selected as the training data from the four domains, respectively. 20 random train/test splits are implemented, and the average recognition accuracies of FSNM [52], FSSI [48], SFSS [51], MLHR [9] and the proposed GLCC are reported in Table VI.

From Table VI, we observe that the object recognition performance is well improved by using the deep representation based on CNN features for all methods. The proposed GLCC still outperforms others, except that the proposed method is 0.1% lower than FSSI for Dslr domain. Note that the objective of the proposed method is for multi-feature learning, while deep feature is recognized as only one kind of feature. Therefore, in the experiment, we consider the outputs of the 6[th] and 7[th] layer of CNN as two kinds of deep features. The results demonstrate the generalization of GLCC as a multi-feature shared learning framework, regardless of the feature types (e.g. conventional descriptors or deep features).

TABLE V
MEANS AND STANDARD DEVIATIONS (%) OF MAPs OVER SIX EVENTS FOR ALL METHODS IN THREE CASES

| Method | FSNM [52] | FSSI [48] | SFSS [51] | MLHR [9] | **GLCC** |
|---|---|---|---|---|---|
| MAP | 7.20±0.20 | 9.03±0.09 | 7.63±0.10 | 8.94±0.09 | **9.36±1.05** |
| Time (s) | 10.1 | 5.6 | 9.8 | 7.4 | 6.2 |

TABLE VI
RECOGNITION ACCURACY ON CNN-FEATURES OF 4 DOMAINS

| Method | FSNM [52] | FSSI [48] | SFSS [51] | MLHR [9] | **GLCC** |
|---|---|---|---|---|---|
| Amazon | 92.2±0.27 | 93.5±0.21 | 92.4±0.16 | 93.3±0.20 | **93.9±0.13** |
| Caltech | 79.9±0.48 | 84.3±0.49 | 81.6±0.31 | 85.0±0.40 | **85.5±0.37** |
| Webcam | 96.2±0.38 | 96.3±0.32 | 96.0±0.31 | 96.0±0.38 | **96.8±0.37** |
| Dslr | 96.9±0.59 | **97.6±0.60** | 97.0±0.37 | 96.8±0.60 | 97.5±0.66 |

### H. Weights of the Laplacian and Hessian Graph

The proposed method uses the group graph regularizer based on the weighted Laplacian and Hessian graphs for semi-supervised multi-feature learning. The learned weights $\boldsymbol{\alpha} = [\alpha_1, \cdots, \alpha_m]$ and $\boldsymbol{\beta} = [\beta_1, \cdots, \beta_m]$ of the Laplacian and Hessian graphs with $m$ features are provided in Table VII. In general, on a particular dataset, each feature should be with different contribution to the recognition. As shown in Table VII, the learned weights $\alpha_i$ ($i = 1, \cdots, m$) of the Laplacian graphs approach the average value $1/m$, that is, the weight is close to 0.14, 0.25, 0.25, and 0.33 for the Flower 17 data, Caltech 101, YouTube & Consumer videos data and NUS-WIDE data, respectively. Instead, the divergence of the learned weight $\beta_i$ ($i = 1, \cdots, m$) of the Hessian graph is more visible, such that the optimal weight for each feature is achieved. Thus, we believe that the group graph with the Laplacian and Hessian graphs may be more flexible and effective in semi-supervised multi-feature learning for pursuit of a robust performance.

TABLE VII
LEARNED WEIGHTS OF THE LAPLACIAN AND HESSIAN GRAPHS FOR DIFFERENT DATASETS

| Dataset | Flower 17 data | | | | | | | Caltech 101 data | | | |
|---|---|---|---|---|---|---|---|---|---|---|---|
| Feature | HOG | HSV | SiftInt | SiftBdy | Color | Shape | Texture | PhowColor | PhowGray | SSIM | GB |
| α | 0.14 | 0.14 | 0.15 | 0.14 | 0.14 | 0.15 | 0.14 | 0.25 | 0.25 | 0.25 | 0.25 |
| β | 0.14 | 0.16 | 0.12 | 0.10 | 0.16 | 0.14 | 0.18 | 0.24 | 0.25 | 0.25 | 0.26 |
| Dataset | YouTube&Consumer video data | | | | | | | Large-scale NUS-WIDE data | | | |
| Feature | SIFT (L=0) | | SIFT (L=1) | | STIP (L=0) | | STIP (L=1) | EDH | | CORR | WT |
| α | 0.26 | | 0.26 | | 0.24 | | 0.24 | 0.32 | | 0.34 | 0.34 |
| β | 0.01 | | 0.15 | | 0.25 | | 0.59 | 0.31 | | 0.39 | 0.30 |

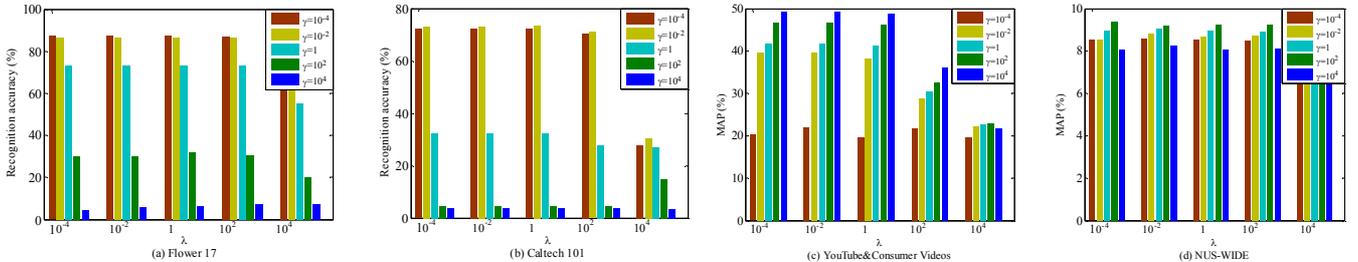

Fig. 4 Performance variation of GLCC w.r.t. λ and γ on Flower 17, Caltech 101, YouTube & Consumer Video and NUS-WIDE datasets

### I. Parameter Sensitivity Analysis

The parameter sensitivity analysis of the two trade-off parameters λ and γ that control the complexity and overfitting of the proposed model is discussed in this section. Specifically, λ and γ are tuned from the set $\{10^{-4}, 10^{-2}, 1, 10^{2}, 10^{4}\}$ in experiments. The performance variations (i.e. recognition accuracy/MAP) with different values of λ and γ for different datasets are described in Fig.4, from which we have the following observations: 1) a small value of each parameter contributes much better performance for the Flower 17 and Caltech 101 data (see Fig.4-a and Fig.4-b). In particular, the performance deteriorates sharply when γ is larger than 1; 2) for the YouTube & Consumer videos (see Fig.4-c), a larger value of γ and a small value of λ are more effective; 3) for the NUS-WIDE (see Fig.4-d), the best result is obtained when γ=100; 4) the parameter λ shows a relatively stable performance for the Flower 17 and NUS-WIDE datasets. Additionally, λ can be set as 1 for all datasets such that only one parameter γ is free and the parameter tuning is easily achieved.

### V. CONVERGENCE AND COMPUTATIONAL TIME ANALYSIS

In this section, the convergence analysis and the computational time of the proposed model on several datasets are discussed.

### A. Convergence Analysis

The convergence proofs of the proposed model are provided in section III (part E). The convergence curves of the proposed objective function (11) over iterations on the four datasets such as Flower 17, Caltech 101, YouTube & Consumer Videos and NUS-WIDE are described in Fig. 5. One can observe that after a few iterations the objective function can converge to a stable value. Additionally, we have also analyzed the convergence of the difference $\Delta P_t = \sum_{i=1}^{m} \|\mathbf{P}_t^i - \mathbf{P}_{t-1}^i\|_F$ in iteration $t$. The curves of $\Delta P_t$ over iterations for the four datasets are described in Fig. 6. It is clearly seen that the difference $\Delta P_t$ for each dataset always converges to a small value after several iterations. The efficiency of the proposed method with fast convergence can be shown.

### B. Computational Time Analysis

The total computational time (s) on the Flower 17, Caltech 101, YouTube & Consumer Videos and NUS-WIDE datasets has been reported in Table II, Table III, Table IV and Table V, respectively. From these tables, we can observe that the proposed method has a comparative computational power. Note that the experiments on the Flower 17, Caltech 101 and YouTube & Consumer video datasets are executed in a laptop with Inter Core i5 CPU (2.50GHz) and 4 GB RAM. The experiment on the NUS-WIDE web image dataset is executed in a computer with Inter Core i7 CPU and 32GB RAM.

### VI. CONCLUSION

In this paper, we propose a multi-feature shared learning framework for visual understanding such as object recognition, video event recognition and image classification. The proposed method is the so-called Global-Label-Consistent classifier (i.e. GLCC), which includes several significant advantages. First, the proposed GLCC makes full consideration of the complementary *structural* information of multiple features for robust recognition performance. Second, motivated by the semi-supervised manifold regression, a group graph manifold regularizer composed of the weighted Laplacian and Hessian graphs of multiple features is proposed for manifold structure preservation of the intrinsic data geometry. For this reason, the global consistency (i.e. the label prediction of each feature is consistent with the global prediction of all features) is well exploited. Third, a $\ell_2$-norm based global classifier with an alternative optimization solver is proposed, such that the model is more computationally efficient. Finally, the model is experimented on various visual benchmark datasets. Comparisons with state-of-the-arts demonstrate that the proposed method is very effective in recognition performance and efficient in computation.

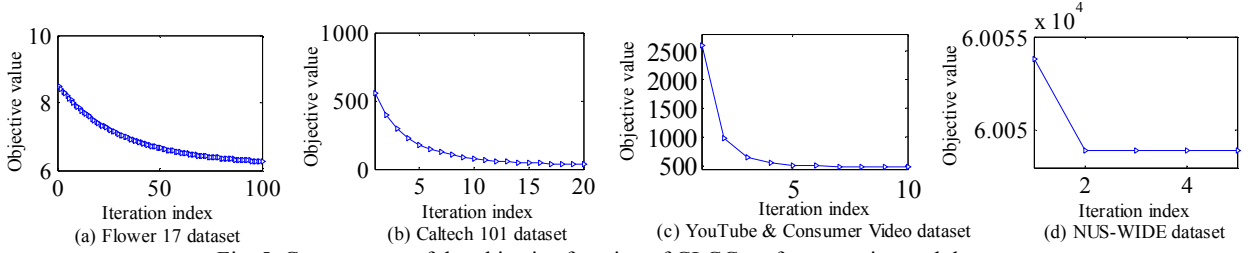

Fig. 5. Convergence of the objective function of GLCC on four experimental datasets.

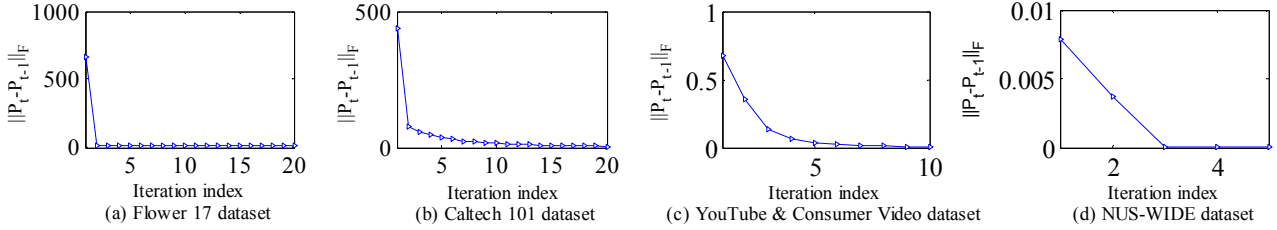

Fig. 6. Covergence of $\Delta P_t$ of GLCC on four experimental datasets, where $t$ denotes the current index of iteration.

In the future work, active learning and selection of the most useful features instead of hand-crafted features would be an interesting topic, particularly for high-dimensional features (e.g. CNN features) in large scale multimedia applications.

## ACKNOWLEDGMENT

We would like to express our sincere appreciation to the Associate Editor and the anonymous experts for their insightful comments, which has greatly improved the quality of the paper. We would also like to thank Dr. Cai and Dr. Kadri for their help in proofreading the paper in language and experiments.

## APPENDIX A

The total Hessian energy estimation of single view/feature can be represented as [5]

$$\hat{S}_{Hess}(f) = \langle F, \Omega F \rangle = \text{Tr}(F^T \Omega F) \quad ①$$

where $\Omega$ is the sparse Hessian energy matrix of the training set.
*Proof:* First, a local tangent space $T_{X_i}M$ of data point $X_i$ is defined. In order to estimate the local tangent space, PCA is performed on the $k$ nearest neighbors space $N_k(X_i)$, then $m$ leading eigenvectors can be obtained as the orthogonal basis of $T_{X_i}M$. The Hessian regularizer defined as $\|\nabla_a \nabla_b f\|^2|_{X_i}$ of the data point $X_i$, is the squared norm of the second covariant derivative, which corresponds to the Frobenius norm of the Hessian of $f$ at the normal coordinates.

$$\|\nabla_a \nabla_b f\|^2|_{X_i} = \sum_{r,s=1}^{m} \left(\frac{\partial^2 f}{\partial x_r \partial x_s}|_{X_i}\right)^2 \quad ②$$

where the partial derivative is computed as

$$\frac{\partial^2 f}{\partial x_r \partial x_s}|_{X_i} \approx \sum_{j=1}^{k} H_{rsj}^{(i)} f(X_j) \quad ③$$

Substitute ③ into ②, the estimation of the Frobenius norm of the Hessian of $f$ at the data point $X_i$ is expressed as

$$\|\nabla_a \nabla_b f\|^2|_{X_i} = \sum_{r,s=1}^{m} \left(\frac{\partial^2 f}{\partial x_r \partial x_s}|_{X_i}\right)^2 = \sum_{r,s=1}^{m} \left(\sum_{\alpha=1}^{k} H_{rs\alpha}^{(i)} f_\alpha\right)^2$$
$$= \sum_{\alpha,\beta=1}^{k} f_\alpha f_\beta \Omega_{\alpha\beta}^{(i)}$$

where $\Omega_{\alpha\beta}^{(i)} = \sum_{r,s=1}^{m} H_{rs\alpha}^{(i)} H_{rs\beta}^{(i)}$.

Then, the total estimated Hessian energy, defined as the sum over all data points, can be represented as

$$\hat{S}_{Hess}(f) = \sum_{i=1}^{n} \|\nabla_a \nabla_b f\|^2|_{X_i} = \sum_{i=1}^{n} \sum_{r,s=1}^{m} \left(\frac{\partial^2 f}{\partial x_r \partial x_s}|_{X_i}\right)^2$$
$$= \sum_{i=1}^{n} \sum_{\alpha \in N_k(X_i)} \sum_{\beta \in N_k(X_i)} f_\alpha f_\beta \Omega_{\alpha\beta}^{(i)} = \langle F, \Omega F \rangle = \text{Tr}(F^T \Omega F)$$

The proof of ① is completed.

## APPENDIX B

To solve the equation group (21) in the paper, we first show the details of solving the $\alpha_i$ as follows.

The first and the third equations in (21) can be combined as

$$\begin{cases} r\alpha_i^{r-1} \text{Tr}(F^T \mathcal{L}^{(i)} F) - \mu = 0 \\ \sum_{i=1}^{m} \alpha_i - 1 = 0 \end{cases} \quad ④$$

For the first equation in ④, there is

$$[r\alpha_i^{r-1} \text{Tr}(F^T \mathcal{L}^{(i)} F)]^{\frac{1}{r-1}} = \mu^{\frac{1}{r-1}}$$
$$\downarrow$$
$$\alpha_i = \mu^{\frac{1}{r-1}} / r^{\frac{1}{r-1}} \left(\text{Tr}(F^T \mathcal{L}^{(i)} F)^{\frac{1}{r-1}}\right) \quad ⑤$$
$$\downarrow$$
$$\sum_{i=1}^{m} \alpha_i = \mu^{\frac{1}{r-1}} / r^{\frac{1}{r-1}} \cdot \sum_{i=1}^{m} \left(1/\text{Tr}(F^T \mathcal{L}^{(i)} F)^{\frac{1}{r-1}}\right) \quad ⑥$$

Consider the $2^{th}$ equation in ④ and the equation ⑥, we have

$$\mu^{\frac{1}{r-1}} / r^{\frac{1}{r-1}} = 1/\sum_{i=1}^{m} \left(1/\text{Tr}(F^T \mathcal{L}^{(i)} F)^{\frac{1}{r-1}}\right) \quad ⑦$$

Substitute ⑦ into ⑤, we can obtain $\alpha_i$ as (22). Similarly, $\beta_i$ can also be solved with the same steps as that of solving $\alpha_i$.